\documentclass[runningheads]{llncs}

\usepackage{graphicx}
\usepackage{comment}
\usepackage{amsmath,amssymb}
\usepackage{color}
\usepackage{url}
\usepackage{hyperref}

\usepackage{graphicx}
\usepackage{amsmath}
 \usepackage{nicematrix} 
\usepackage{multirow}
\usepackage{comment}
\usepackage{float} 

\usepackage{tikz, pgfplots, fix-cm}
\usetikzlibrary{plotmarks}
\usepackage{graphicx,booktabs,array}

\usepackage{amssymb}
\usepackage{pifont}
\newif\ifreview
\reviewfalse

\ifreview
	\usepackage{lineno}

	\linenumbers
\fi

\def\secref#1{Sec.~\ref{#1}}
\def\figref#1{Fig.~\ref{#1}}
\def\tabref#1{Tab.~\ref{#1}}
\def\eqref#1{Eq.~(\ref{#1})}

\newcommand\etal{\emph{et al.}}

\begin{document}

\def\SubNumber{45}

\def\GCPRTrack{Regular Track}

\title{Virtual Temporal Samples for Recurrent Neural Networks: applied to semantic segmentation in agriculture}

\ifreview
\else

	\author{Alireza Ahmadi\orcidID{0000-0001-7909-094X} \and
	Michael Halstead\orcidID{0000-0001-7185-9304} \and
	Chris McCool\orcidID{0000-0002-0577-1299}}
	
	
	\authorrunning{F. Author et al.}
	
	\institute{University of Bonn, Nussallee 5, Bonn 53115 Germany \\
	\url{http://agrobotics.uni-bonn.de}\\
	\email{\{alireza.ahmadi, michael.halstead, cmccool\}@uni-bonn.de}}
\fi

\maketitle              

\thispagestyle{empty}
\pagestyle{empty}

\begin{abstract}



This paper explores the potential for performing temporal semantic segmentation in the context of agricultural robotics without temporally labelled data.
We achieve this by proposing to generate \textit{virtual} temporal samples from labelled still images.
By exploiting the relatively static scene and assuming that the robot (camera) moves we are able to generate virtually labelled temporal sequences with no extra annotation effort.
Normally, to train a recurrent neural network (RNN), labelled samples from a video (temporal) sequence are required which is laborious and has stymied work in this direction.
By generating virtual temporal samples, we demonstrate that it is possible to train a lightweight RNN to perform semantic segmentation on two challenging agricultural datasets.
Our results show that by training a temporal semantic segmenter using virtual samples we can increase the performance by an absolute amount of $4.6$ and $4.9$ on sweet pepper and sugar beet datasets, respectively.
This indicates that our \textit{virtual} data augmentation technique is able to accurately classify agricultural images temporally without the use of complicated synthetic data generation techniques nor with the overhead of labelling large amounts of temporal sequences.

\textit{Keywords}—temporal data augmentation; spatio-temporal segmentation; agricultural robotics.

\end{abstract}

\section{Introduction}
\label{sec:introduction}

In recent years, agricultural robotics has received considerable attention from the computer vision, robotics, and machine learning communities due, in part, to its impact on the broader society. 
Agricultural applications, such as weeding and harvesting are demanding more automation due to the ever increasing need for both quality and quantity of crops. 
This increased attention has partly driven the development of agricultural platforms and expanded their capabilities, including quality detection \cite{halstead2018fruit}, autonomous weeding~\cite{slaughter2008autonomous} and automated crop harvesting~\cite{lehnert2016sweet}.

An advantage of surveying an agricultural scene over a pedestrian scene is the structured and relatively static nature of crops, particularly with respect to the moving platform.
While the scene is somewhat static and structured there are a number of challenges an automatic agent needs to overcome due to the complicated nature of the scene, including, illumination variation, and occlusion.
Recent advances in agricultural robotics have exploited convolutional neural networks (CNNs)~\cite{krizhevsky2012imagenet} to both alleviate some of these challenges and achieve high performance.

Modern machine learning techniques rely on CNNs to perceive useful visual information about a scene. 
Most of the successful deep learning techniques utilize a paradigm of multi-layer representation learning from which semantic segmentation~\cite{krizhevsky2012imagenet}~\cite{lecun1998gradient} has evolved. 
These segmentation networks are able to classify on a pixel level~\cite{long2015fully} the appearance of a specific class, creating class based output maps. 
From a spatio-temporal perspective RNNs~\cite{zhang2018spatial} are able to exploit previous information to improve performance in the current frame.
Despite these advances, only feed-forward networks have been predominately used to generate the network parameters in each layer~\cite{fawakherji2019crop}.
In Agriculture one of the few examples which uses spatio-temporal information is \cite{lottes2018fully} where they exploit the regular planting intervals of crop rows. However, this approach is still ill suited to more generalised agricultural environments without such structure like fruit segmentation in horticulture.
In this paper, we implement a lightweight RNN based on the UNet~\cite{ronneberger2015u} architecture by employing ``feedback'' in the decoder layers.
These ``feedback'' layers are used to perceive spatio-temporal information in an effort to improve segmentation accuracy.

\begin{figure}[t]
	\centering
	\includegraphics[width=0.8\linewidth]{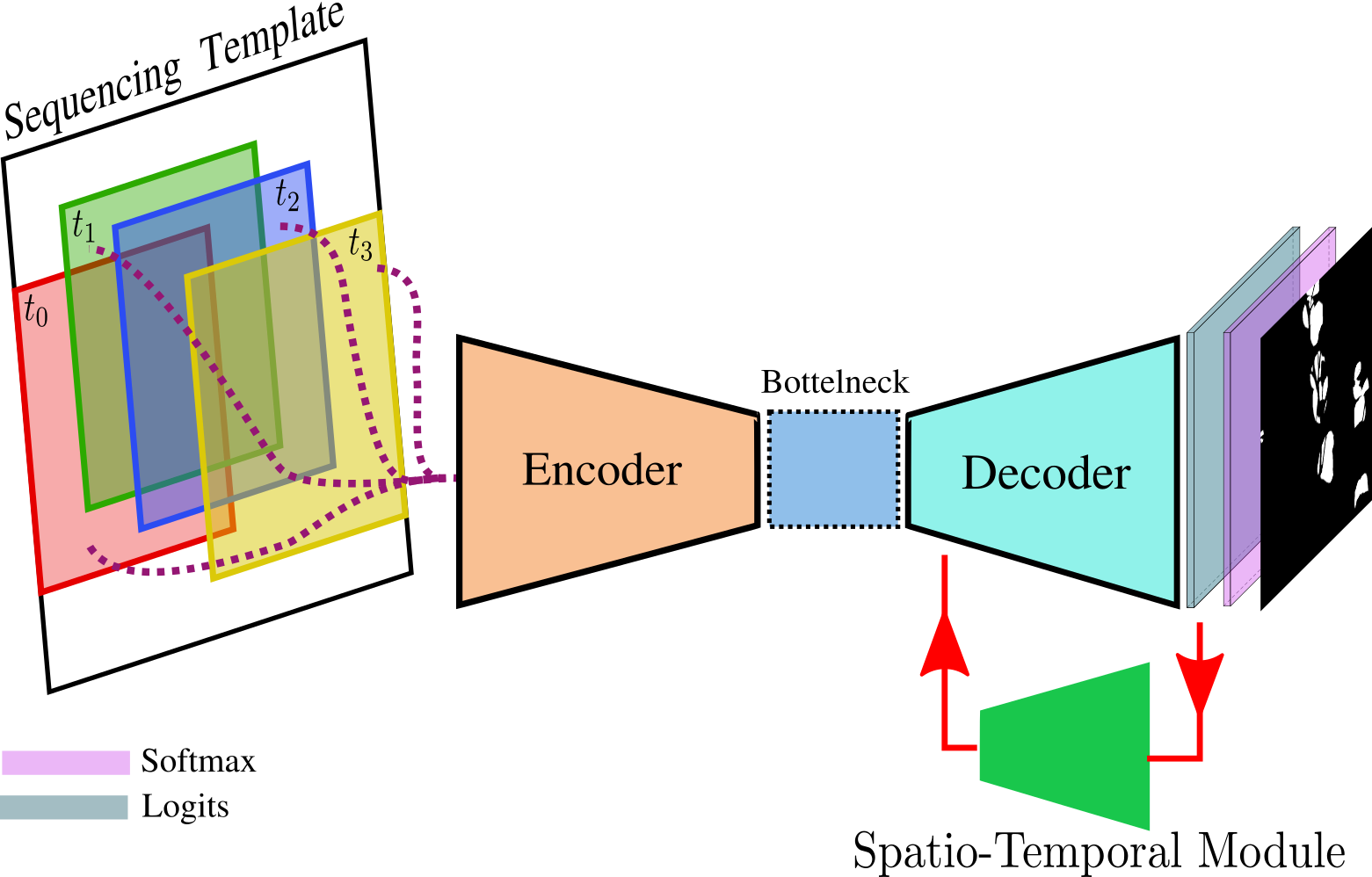}
	\caption{Overall structure of our proposed spatio-temporal network. The input to this network is being drawn out of the sequencing template, considering the order of virtual sequences based on direction of motion and time. The spatio-temporal module feeds back feature maps to network and forces network to learn frames dependency.}
	\label{fig:motivation}
\end{figure}

We demonstrate that it is possible to train an RNN by generating virtual temporal sequences from annotated still images.
This is important as the majority of agricultural datasets either do not contain temporal information or perform sparse labelling of the frames (consecutive image frames are rarely annotated). 
While creating annotations between the labelled frames can be achieved through weakly labelled techniques~\cite{shi2017weakly}, these approaches are often noisy and introduce unwanted artifacts in the data.
We are able to generate virtual temporal sequences by augmenting the annotated still images via successive crops and shifts inside the original annotated image as outlined in~\figref{fig:motivation}.
By utilising a crop and shift method of augmentation we are able to maintain the structural information of the scene, similar to traditional temporal sequences.
The validity of this virtual temporal sequence approach and the RNN structure are evaluated using two annotated agricultural datasets which represent vastly different scenes (field vs glasshouse), motions, crops, and camera orientations; see Section~\ref{subsec:bupseq}.

In this work the following contributions are made:
\begin{enumerate}
  \item We propose a method for generating virtual temporal sequences from a single annotated image, creating spatio-temporal information for training an RNN model;
%
  \item We explore different methods to perceive spatio-temporal information in an RNN-UNet structure and show that a convolution based module outperforms feature map downsizing through bi-linear interpolation;

  \item Our proposed lightweight RNN architecture along with our novel data augmentation technique
  is able to improve semantic segmentation performance on different datasets regardless of the distribution of objects in the scene and considerably different nature of the scenes.
%
\end{enumerate}

This paper is organised in the following manner:~\secref{sec:relatedworks} reviews the prior work;~\secref{sec:dataset} discusses our proposed virtual samples generation method; the proposed temporal approach is explained in~\secref{sec:temporal_approach};~\secref{sec:exp} describes our experimental setup and implementation details; the results are detailed in~\secref{sec:results};~and finally the conclusions \secref{sec:conc}.

\section{Related Works}
\label{sec:relatedworks}

Recently, the problem of semantic segmentation has been addressed by a number of different approaches including~\cite{ronneberger2015u,Noh_2015_ICCV,wang2018two}.
Most of the state-of-the-art approaches used fully convolutional networks (FCNs)~\cite{long2015fully} as an end-to-end trainable, pixel-level classifier~\cite{simonyan2014very}. 
However, in an agricultural robotic setting a number of extra challenging factors need to be overcome including: non-static environments, highly complex scenes and significant variation in illumination.

In an effort to circumvent  these factors researchers have enhanced standard semantic segmentation by embedding spatio-temporal information~\cite{wang1994spatio,He_2017_CVPR} into their architectures.
In these cases, where spatio-temporal information was available, integrating this information improved performance~\cite{campos2016spatio,lottes2018fully}.
Furthermore, Jarvers and Neumann~\cite{jarvers2019incorporating} found that by incorporating sequential information, errors which occur in one frame could be recovered in subsequent frames.

In an agricultural context, Lottes~\etal\ \cite{lottes2018fully} improved their semantic segmentation network performance by incorporating spatial information about the sequence of images. 
By exploiting the crop arrangement information (geometric pattern of sowing plants in the field) they improved segmentation performances. 
While this created promising results in field settings the geometric pattern assumption does not hold in agriculture, consider fruit segmentation in horticulture.

In \cite{zhang2018spatial}, the authors proposed RNNs as a method of reliably and flexibly incorporating spatio-temporal information. 
A key benefit witnessed by most spatio-temporal techniques was the layered structure of FCNs, which provide the opportunity to embed ``feedback'' layers in the network.

By embedding these ``feedback'' layers researchers can make use of the extra content and context provided by the multiple views (observations) of the same scene.
This embedding acts on the system by biasing future outputs at $t+1$ based on the current output at $t$.
Benefits of this method were found to hold when spatio-temporal dependencies exist between consecutive frames~\cite{tsuda2020feedback}.
By comparing feed-forward and feed-backward networks~\cite{jarvers2019incorporating} showed the later was able to increase the receptive fields of the layers.
Ultimately, ``feedback'' layers provide richer feature maps to enable RNN-based systems to improve predictions.

Long Short-Term Memory (LSTM)~\cite{hochreiter1997long} based architectures have also been shown to improve classification tasks by integrating temporal information~\cite{pfeuffer2019semantic,cordts2016cityscapes}.
These LSTM cells can be inserted into the network to augment performance, however, they add significant complexity to the network.

Another method used to improve the generalisability of networks is data augmentation.
Kamilaris et al.~\cite{kamilaris2018deep} provided a comprehensive survey of early forms of data augmentation used by deep learning techniques.
Generally, these methods include rotations, partitioning or cropping, scaling, transposing, channel jittering, and flipping~\cite{yu2017deep}.
Recent advances in data augmentation have lead to generative adversarial networks (GANs) which have the ability to generate synthetic data~\cite{douarre2019novel}, enhancing the generalisability of the trained models.
However, this adds further complexity to the pipeline as generating synthetic data can often be a time consuming exercise.



Generally, recent advances in machine learning have been made by using large labelled datasets such as ImageNet~\cite{deng2009imagenet} which was used for object classification.
These datasets only exist as labour is directed towards the particular task.
This provides a major hurdle for deploying temporal approaches to novel domains such as agriculture.

To overcome the data requirements of spatio-temporal techniques, we make use of the partially labelled data of~\cite{smitt2020pathobot} and a newly captured dataset of sugar beets.
Our proposed temporal data augmentation approach (see ~\secref{sec:dataset}) generates virtual samples that only represent short-term temporal information, as such we do not explore LSTM-based approaches.
This consists of dense (small spatial shifts between frames) temporal sequences that we augment and use to train a lightweight RNN architecture.

\section{Generating Virtual Spatio-Temporal Sequences}
\label{sec:dataset}

\begin{figure*}[t!]
	\vspace{2mm}
	\centering
	\includegraphics[width=0.98\linewidth, keepaspectratio]{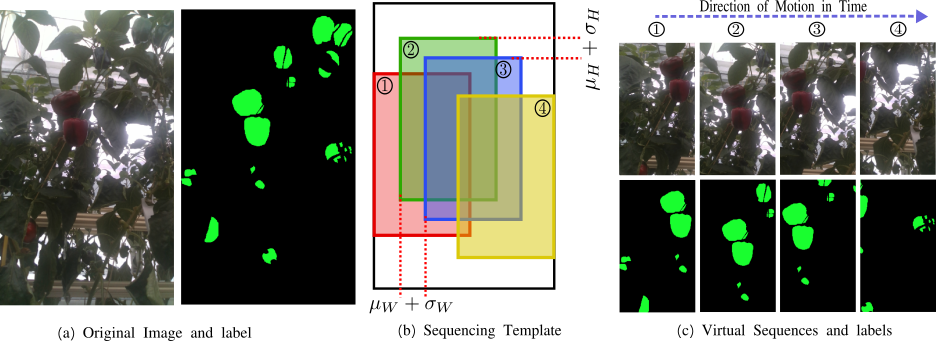}
	\caption{An example virtual sequence generation. (a) Original image and label, (b) Sequencing template used for generating virtual frames with $N=4$. Frame one is cropped at the position that will ensure N frames can be created without going beyond the border of the image, considering the means \(\mu_W , \mu_H\) and standard deviations \(\sigma_W , \sigma_H\), (c) Generated virtual sequence with labels.}
	\label{fig:vs_scheme}
\end{figure*}
We propose that \textit{virtual} sequences can be generated from an existing image and its annotation.
The virtual sequences can simulate camera motion by performing consecutive crops and shifts as shown in~\figref{fig:vs_scheme}.
This has the advantage that the virtual sequence is fully annotated (per pixel) without having to perform laborious annotation or risk the propagation of noisy labels.
While this approach is a simulation of camera motion and removes the natural occlusion witnessed in actual motion, it should reduce the requirement for mass annotation of large-scale datasets and provides fully labelled spatio-temporal data.
Structuring motion in this manner (both the $x$ and $y$ direction) is important for simulating traditional temporal data on a robot where limited rotation occurs, for this reason we avoid other augmentation techniques such as up-down flipping and affine rotations.
The aim of this data augmentation technique is to provide extra contextual information which enables a network to learn the relationship between consecutive frames and improve prediction accuracy.
A limitation of this approach is that it assumes that the robot (camera) moves and that the scene is relatively static.


Virtual sequences are generated by employing a crop and shift technique based on manually  obtained parameters.
There are two sets of parameters that are needed.
First, we need to estimate the movement parameters in the actual data (in image coordinate frames).
Second, we need to define the number of frames $N$ that will be generated for each \textit{virtual} sequence.
The number of frames $N$, sets the number of \textit{virtual} samples to be simulated per image in the dataset, see~\figref{fig:vs_scheme}. 
Once $N$ is known we replicate the natural motion of the camera in the scene by cropping $N$ fixed sized cropped images from the original image. 
Using the two sets of parameters, each cropped frame is then computed by,

\begin{equation}
    \begin{split}
    	W_n & = W_s + \mathcal{N}\left( \mu_w, \sigma_w^{2} \right) \lambda n, \\
    	H_n & = H_s + \mathcal{N}\left( \mu_h, \sigma_h^{2} \right) \gamma n.
    \end{split}
\end{equation}
\noindent $H_n$, $W_n$ denotes the new position of top-left corner of the cropped image in the original image, $W_s, H_s$ represent the start position, and $n$ is the frame index ranging from $0$ to $N-1$. 
The ``directions of travel'' $\lambda$ and $\gamma$ are chosen between left-to-right ($\lambda=1$), up-to-down ($\gamma=1$) and right-to-left ($\lambda=-1$), down-to-up ($\gamma=-1$) for each annotated sample and are fixed for all frames within the sequence. 
This ensures no bias is introduced while training, considering the nature of motion in real data (platforms move in one direction). 
The directional bias is essential as the camera motion occurs in both directions for the BUP20 dataset (main motion in $x$ direction), while SB20 only captured up-to-down motions (along the $y$ axis). 

\section{Proposed Temporal Network}
\label{sec:temporal_approach}

\begin{figure}[t!]
	\centering
    \includegraphics[width=1.0\linewidth]{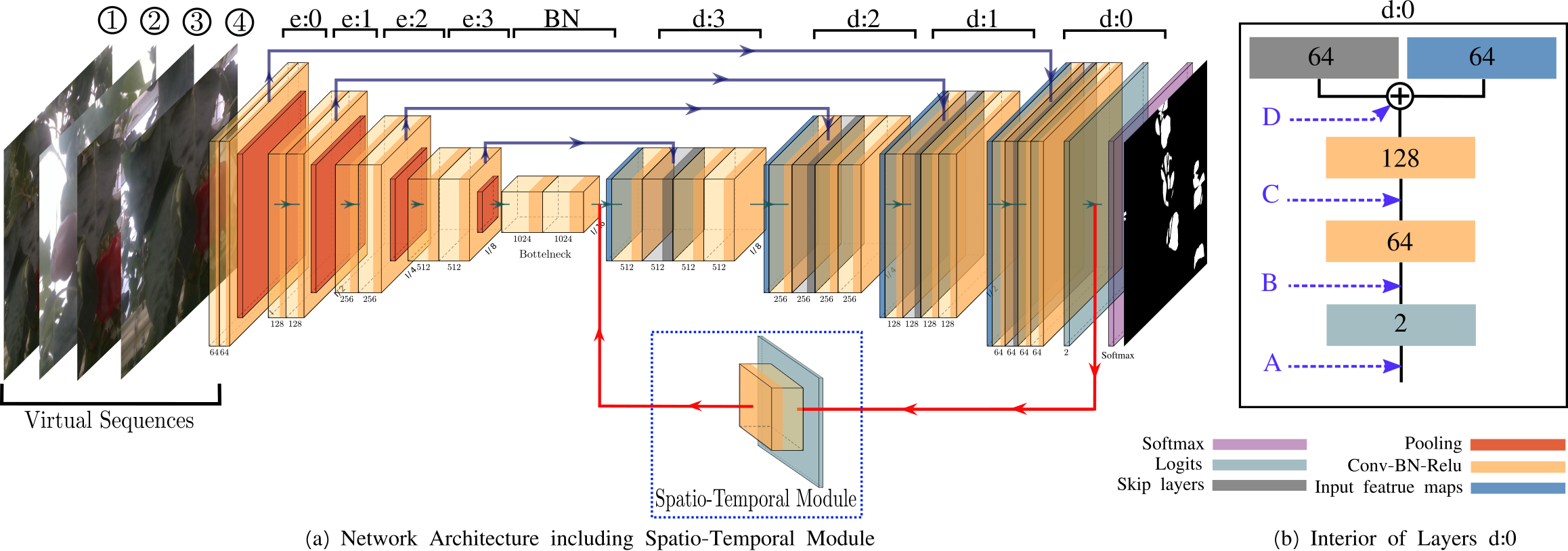}
    \caption{Network architecture. At the input, four sequences of virtually generated frames of a single image are passed to the network, separately. In each pass only one of the samples will be fed to the network (batch size = 1). The spatio-temporal module is feeding back 2 last layers (before the soft-max) to the layer \textit{d:3-D}}
	\label{fig:net_arch}
	\vspace{-6mm}
\end{figure} 

The primary research goal of this paper is to explore spatio-temporal relations of a CNN without the need for laborious manual labelling of temporal sequences. As such, we use a lightweight neural network structure based on the UNet \cite{ronneberger2015u} architecture; we also use UNet to produce baseline results for comparison.


From a spatio-temporal standpoint we implement feedback modules within the UNet architecture.
We augment the UNet baseline architecture to allow different types of feedback within its current composition of layers.
\figref{fig:net_arch}-a outlines both the baseline system and the inclusion of the spatio-temporal module. 
While a number of techniques exist to provide feedback~\cite{jarvers2019incorporating}, we use layer concatenation to join the feedback with the feed forward channels.
Concatenation was selected as it directly adds information to the feature map of the host layer, creating greater potential to learn direct relationships between the current and previous layers (frames).

In~\figref{fig:net_arch}-a, all layers are assigned a specific name and number. 
Specifically, the $n$-th layer of the decoder ($d$) and encoder ($e$) are denoted to as \textit{d:n} and \textit{e:n} respectively and the bottleneck is denoted as $BN$.
We access the sub-layers within the main layers of the network via the intervention points assigned with names ranging from $A$ to $D$. 
For instance, layer \textit{d:0} with its sub-divisions is depicted in  \figref{fig:net_arch}-b, containing four intervention points. 
Each intervention point can be used as an extraction or insertion point for the spatio-temporal module.

A key complexity when using the different layers as a feedback to the network is the discrepancy between the two resolutions. 
The disparity between layer height and width when feeding back creates additional complexity and needs to be allowed for.
For instance, in~\figref{fig:net_arch}-a the feature maps at the sampling layer \textit{d:0-A} with size of $2\times H\times W $ are fed back to the first deconvolution layer after the bottleneck in \textit{d:3-D} of the baseline architecture with size of $1024\times H/16\times W/16$. 
To alleviate this issue we propose two methods of modifying the feedback layer: 1) bi-linear interpolation (\textit{Bi-linear}); and 2) $2D$ convolution re-sampling (\textit{Conv}).
The \textit{Bi-linear} approach maintains the integrity of the feature map such that the re-sampled output is similar to the input; just down or up-sampled. 
By contrast, the \textit{Conv} re-sampling adds the benefit of learning to re-sample as well as learning to transform the feature map.
A limitation of the \textit{Conv} approach is the added depth (complexity) required to produce the feedback feature map.
The (\textit{Conv}) re-sampling block consists of a 2D convolutional \textit{block} with a $3\times3$ kernel (of stride $2$ and padding $1$) followed by batch normalization and a \textit{ReLU} activation function.
In training and evaluation phases, at the start of a new temporal (\textit{real} or \textit{virtual}) sequence, the activation values of the spatio-temporal module are set back to $\mathbf{1}$ to avoid accumulating irrelevant information between non-overlapping frames.

\section{Experimental Setup}
\label{sec:exp}

We evaluate our proposed approach on two challenging agricultural datasets.
The two datasets represent two contrasting scenarios, the first is a glasshouse environment with sweet pepper and the second is an arable farm with sugar beet.
In both cases, the data was captured using robotic systems which allowed us to extract estimates of the motion information, we describe each dataset in more detail below.

When presenting results we compare algorithms with and without temporal information.
First, we evaluate the impact of using the generated virtual sequences for still image segmentation. 
We use $N$ virtual samples (per image) to train a non-recurrent network which we compare to a system trained on the full size images (baseline).
This allows us to understand if increasing the number of samples at training, data augmentation using the $N$ virtual samples, is leading to improved performance.
Once again, we only consider crop and shift augmentation so we can directly compare the non-recurrent and recurrent models directly.
Second, we evaluate the proposed \textit{spatio-temporal} system using different feedback points, with a variable number of virtual samples $N$ as well as against a classic RNN system where the last frame in the sequence is the only frame with annotations; only the output from the last frame has a label which can be used to produce a loss. 

\subsection{Datasets}
\label{subsec:bupseq}
The two datasets that we use contain video sequences captured from sweet pepper (BUP20) and sugar beet (SB20) fields respectively.
Below we briefly describe each dataset.

\textbf{BUP20} was captured under similar conditions to \cite{Halstead20_1:conference} and first presented in~\cite{smitt2020pathobot}, however, it was gathered with a robot phenotyping platform.
It contains two sweet pepper cultivar, \textit{Mavera} (yellow) and \textit{Allrounder} (red), and was recorded in a glasshouse environment. The dataset is captured from 6 rows,  using 3 Intel RealSense D435i cameras recording RGB-D images as well as IMU and wheel odometry. 
Sweet pepper is an interesting yet challenging domain due to a number of facets. 
Two challenges when segmenting fruit are occlusions caused by leaves and other fruit, and the similarity between juvenile pepper and leaves. 
A sample image is shown in \figref{fig:vs_scheme}.

\begin{figure}[t!]
	\centering
	\begin{tabular}{cc}
        \includegraphics[width=0.33\linewidth]{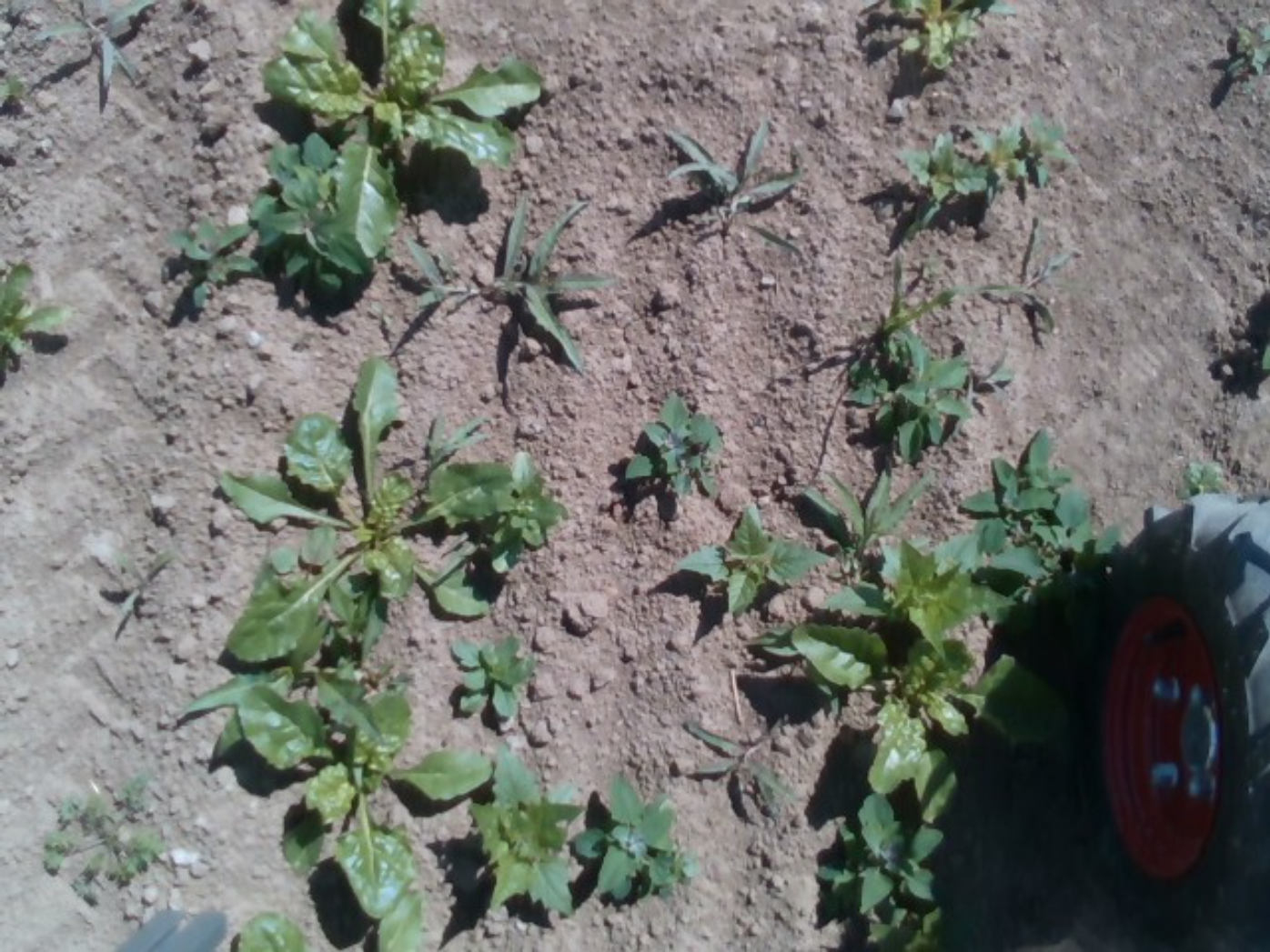}
        \includegraphics[width=0.33\linewidth]{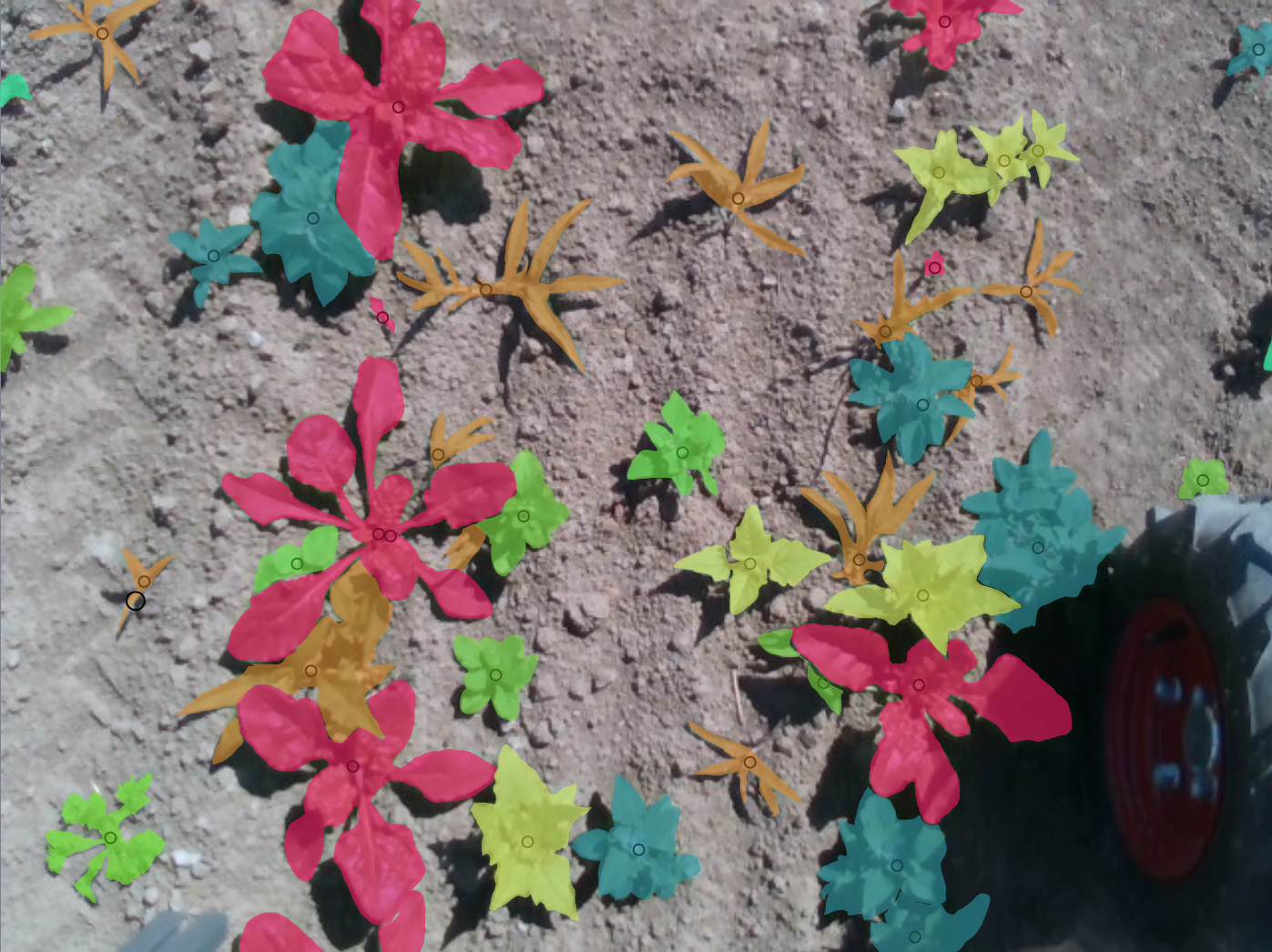}
        \includegraphics[width=0.33\linewidth]{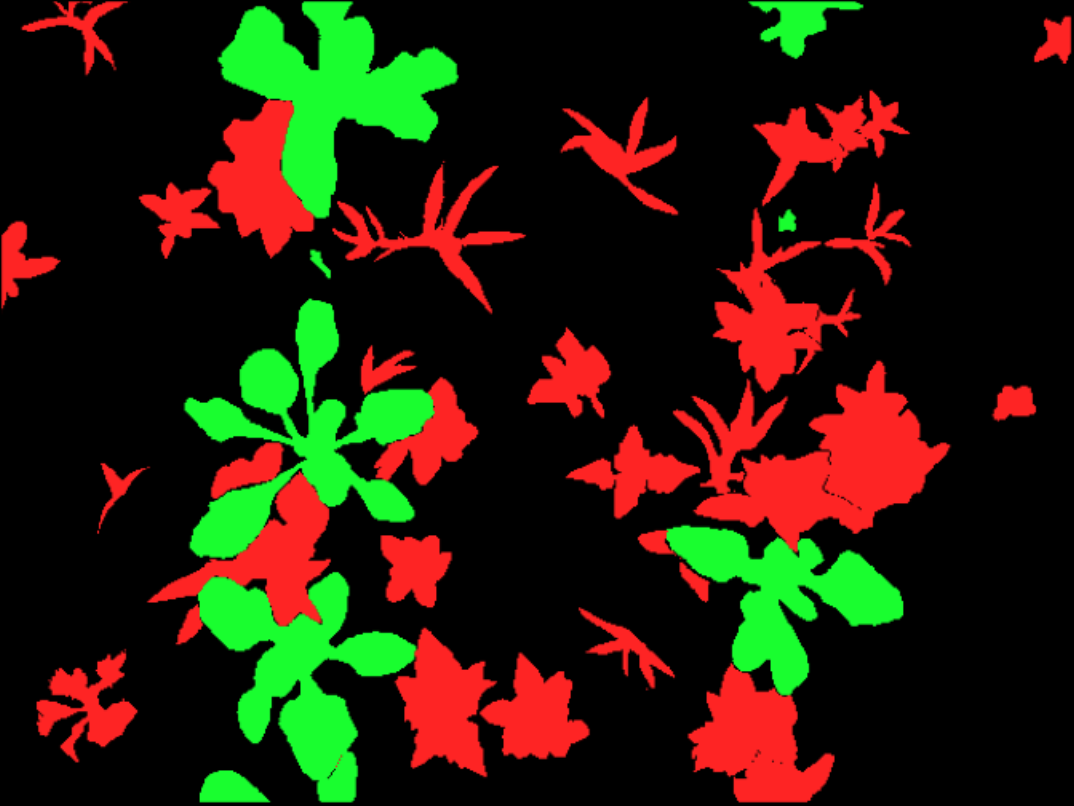}
    \end{tabular}
    \caption{Example image of dataset SB20 (left), same image with multi-class annotations representing different types of crops and weeds using different colors (middle), and (right) shows the crop weed annotation abstraction used in this paper.}
	\label{fig:dataset_sample}
	\vspace{-4mm}
\end{figure}

\begin{table}[!b]
    \vspace{-2mm}
	\centering
	\caption{Dataset Characteristics such as image size, frames per second (fps), number of images for the training, validation and evaluation sets as well as the estimated motion parameters (\textbf{$\mu_{w}$}, \textbf{$\sigma_{w}$}, \textbf{$\mu_{h}$}, \textbf{$\sigma_{h}$}).}
	\begin{tabular}{l l ccc cc cc cc}
	\toprule
	  &          & \textbf{Image Size}  & \textbf{fps} & \textbf{Train} & \textbf{Valid.} & \textbf{Eval.} & \textbf{$\mu_{w}$} & \textbf{$\sigma_{w}$} & \textbf{$\mu_{h}$} & \textbf{$\sigma_{h}$} \\\hline
	 \midrule
    1 & BUP20    & $1280\times720$ & 15 & 124 & 62 & 93 & 5  & 10 & 1 & 3 \\
    2 & SB20     & $640\times480$  & 15 & 71  & 37 & 35 & 2  & 3  & 5 & 7 \\
    \bottomrule
    \end{tabular}
    \vspace{-4mm}
	\label{tab:movement_params}
\end{table}

\textbf{SB20} is a sugar beet dataset that was captured at a field in campus Klein-Altendorf of the University of Bonn.
The data was captured by mounting Intel RealSense D435i sensor with a nadir view of the ground on a Saga robot \cite{grimstad2017thorvald}. 
It contains RGB-D images of crops (sugar beet) and eight different categories of weeds covering a range of growth stages, natural world illumination conditions, and challenging occlusions. 
The captured field is separated into three herbicide treatment regimes ($30\%$, $70\%$, $100\%$) which provides a large range of distribution variations of different classes. 
We only use super-categories: crop, weed, and background for semantic segmentation purposes while, the dataset provides multi-class annotations, the \figref{fig:dataset_sample} shows an example annotated image of this dataset.

Both datasets consist of temporally sparse annotations, that is, the annotation of one image does not overlap with another image.
As such, we can use this data to generate real temporal sequences, of arbitrary size $N$, where \textbf{only} the final frame in the sequence is labelled.
This serves as the real-world temporal data used in our evaluations.
The advantage of creating temporal sequences in this manner is that it includes natural occlusions, varying illumination, and motion blur.
The disadvantage in this method over traditional sequences is that as $N$ increases the total loss decreases by a factor of $1/N$.
This is due to the impact of having only a single annotated spatio-temporal frame to calculate the loss compared to $N$ annotated frames (i.e. the final frame compared to $N$ frames).

\tabref{tab:movement_params} summarizes the information about the images sets provided by BUP20 and SB20 datasets. 
To derive the movement parameters of the \textit{real} sequences, we used odometry data provided from the datasets. 
The estimated parameters express motion of the platforms in both the $x$ and $y$ directions in image coordinate frames, which are summarised in \tabref{tab:movement_params}.

To facilitate the interpretation of our experimental results, we set certain hyper-parameters as constant.
The \textit{real} images of BUP20 dataset are of resized to $704 \times 416$ and its \textit{virtual} images  are generated with a of size $544 \times 320$, also we use SB20 images without resizing and make \textit{virtual} sequences of size $544 \times 416$.
The \textit{still} image experiments are denoted by \textit{\textbf{Still-*}} and use the UNet model. The spatio-temporal (sequence) experiments are denoted by \textit{\textbf{Temporal-*}} and use the lightweight RNN model described in Section~\ref{sec:temporal_approach}, which is based on the UNet model.

\subsection{Implementation and Metrics}
\label{subsec:hyp_exp}

Our semantic segmentation network is based on the UNet model implemented in PyTorch.
To train our network we use Adam~\cite{kingma2014adam} with a momentum of $0.8$ and StepLR leaning rate scheduler with a decay rate of $\gamma=0.8$, decreasing the learning rate every one hundred epochs, starting initially with a learning rate of $0.001$. 
We train all models for a total of 500 epochs with a batch size of $B=4$ using cross-entropy as our loss. 
Also, all models' weights are initialized before training with Xavier method \cite{glorot2010understanding}.
For parameter selection we employ the validation set of the datasets and use the weighted mean intersection over union (mIoU) to evaluate the performance.
For evaluation we also employ the mIoU metric as it provides the best metric for system performance~\cite{rahman2016optimizing}.

\section{Results and Evaluations}
\label{sec:results}


Here we present three studies.
First, we evaluate the performance of still image systems and explore the impact that generating $N$ cropped images, from a single image, has on system performance (data augmentation).
Second, we evaluate the performance of spatio-temporal models when trained using either \textit{real} or \textit{virtually} generated sequences.
Third, we perform an ablation study to explore the impact of varying the number of frames $N$ in the spatio-temporal sequences.
In all cases, we utilize the validation set to select the best performing model.
This can be achieved earlier than the maximum epoch value. 
All of the results that we present are on the \textit{evaluation} set of \textit{real} data (either still images or sequences), unless otherwise stated.

\subsection{Still Image Systems}
\label{subsec:baseline_res}
 
These experiments provide the baseline performance when using only \textit{still} images to train the segmentation system. 
This allows us to investigate if the extra samples generated in the \textit{virtual} sequences lead to improved performance through data augmentation.
For the \textit{Still-Real} experiments only the annotated images were used for training. 
The \textit{Still-Virtual} experiment uses \textit{virtual} samples with $N=5$ (data augmentation) to train the same base-line network.
The results in \tabref{tab:baseline_result} highlight the benefits of the data augmentation in \textit{Still-Virtual} over the \textit{Still-Real}.
There is an absolute performance improvement from \textit{Still-Real} to \textit{Still-Virtual} of $1.4$ and $2.2$ respectively for datasets BUP20 and SB20, which represents a relatively high improvement on the same network gained only through data augmentation.

\begin{table}[t!]
    \vspace{2mm}
	\centering
	\caption{The result of the UNet model when trained and evaluated using \textit{still} images. }
	\begin{tabular}{l l c c}
	\toprule
	  &               & \textbf{BUP20} & \textbf{SB20}  \\\hline
	  \midrule
    1 & \textit{Still-Real}    &  77.3  & 71.3 \\
    2 & \textit{Still-Virtual} &  78.7  & 73.5 \\
    \bottomrule
    \end{tabular}
    \vspace{-4mm}
	\label{tab:baseline_result}
\end{table}

\subsection{Spatio-Temporal Systems}
\label{subsec:st_res}

In these experiments we explore the performance implications of varying how the feedback is provided to our temporal model, as described in Section~\ref{sec:temporal_approach}, and the utility of using \textit{virtual} sequences to train the spatio-temporal model.

First, we investigate the impact of different insertion and extraction points for the spatio-temporal (RNN) model as well as the impact of using either the \textit{Bi-linear} or \textit{Conv} re-sampling methods. 
\tabref{tab:conv_vs_bi} summarises the results of how the feedback is provided to our spatio-temporal model\footnote{Initial empirical results found that insertion and extraction points in the encoder part of the network provided poor performance and so was not considered in further experiments.}.

\begin{table}[h!]
    \vspace{-2mm}
	\centering
	\caption{The outlines of different RNN models using feedback layers extracted from \textit{d:0-B} and looped back to various layers; using Bi-linear  and Covolutional down-sampling feedbacks. Outlines illustrate the performance of different networks trained with \textit{virtual} sequences and evaluated on \textit{real} samples.}
	\begin{tabular}{c l l  c c}
	\toprule
	
	&  & \textbf{Feedback}& \textbf{Conv}  & \textbf{Bi-linear} \\\hline
	\midrule
	\vspace{-3mm}
	\parbox[t]{2mm}{\multirow{5}{*}{\rotatebox[origin=c]{90}{BUP20}}} &&&&\\
	
    & 1 & d:0-B - d:0-D    & 81.9 & 78.9 \\
    & 2 & d:0-B - d:1-D    & 80.3 & 79.1 \\
    & 3 & d:0-B - d:2-D    & 81.1 & 78.9 \\
    & 4 & d:0-B - d:3-D    & 80.8 & 80.4 \\ 
    & 5 & d:0-B - BN       & 80.4 & 78.9 \\ 
    
    \bottomrule
    \vspace{-2mm}
    \parbox[t]{2mm}{\multirow{5}{*}{\rotatebox[origin=c]{90}{SB20}}} &&&&\\ 
    
    & 6  & d:0-B - d:0-D    & 75.4 & 73.9 \\
    & 7  & d:0-B - d:1-D    & 76.2 & 75.1 \\
    & 8  & d:0-B - d:2-D    & 74.5 & 74.3 \\
    & 9  & d:0-B - d:3-D    & 76.0 & 75.5 \\
    & 10 & d:0-B - BN       & 75.5 & 75.1 \\
    
    \bottomrule
    \end{tabular}
    \vspace{-4mm}
	\label{tab:conv_vs_bi}
	
\end{table}



From \tabref{tab:conv_vs_bi} two details are clear, first, \textit{Conv} consistently outperforms its \textit{Bi-linear} counterpart, and second BUP20 results are higher than SB20 across the board.
From the \textit{Conv} versus \textit{Bi-linear} perspective this shows that a learned representation as feedback is of greater benefit than a direct representation.
Then the performance mismatch between the two feedback types also results in different layers performing best, in BUP20 it is \textit{d:0-B-d:0-D} and \textit{d:0-B-d:1-D} for SB20 (lines 1 and 7 respectively of \tabref{tab:conv_vs_bi}).
The variation in performance between the BUP20 and SB20 datasets can be attributed to two key factors.
First, the resolution of SB20 is approximately half of the BUP20 dataset, meaning there is more fine-grained information available in BUP20.
Second, SB20 has more diversity in the sub-classes: the multitude of weed types look significantly different particularly with respect to the crop, while BUP20 only varies the color of a sweet pepper (similarity between sub-classes).
From hereon, all spatio-temporal models use these best performing architectures (see \tabref{tab:conv_vs_bi} - line 1 BUP20, line 7 SB20).



\begin{table}[b!]
	\centering
	\caption{Outlines the results of spatio-temporal models trained with \textit{real} datasets, \textit{virtual} and along with the models train with \textit{virtual} sequences and fine-tuned with \textit{real} temporal frames (line 3).}
	\begin{tabular}{l l c  c}
	\toprule
	  & \textbf{Model}              & \textbf{BUP20} & \textbf{SB20}  \\\hline
	  \midrule 
      1 & \textit{Temporal-Real}               & 78.2 & 73.9 \\
    2 & \textit{Temporal-Virtual}            & 81.9 & 76.2 \\
    3 & \textit{Fine-Tuned}                  & 82.8 & 76.4 \\
    \bottomrule
    \end{tabular}
	\label{tab:st_result}
	\vspace{-4mm}
\end{table}

In~\tabref{tab:st_result} we present the results for several systems trained on \textit{virtual} and \textit{real} sequences, see Section~\ref{sec:temporal_approach} for more details on each system.
It can be seen that all of the spatio-temporal systems outperform the still image systems.
The worst performing system uses only \textit{real} sequences for training and achieves a performance of $78.2\%$ and $73.9\%$, which yield an absolute improvement of $0.9$ and $2.6$ over the \textit{Still-Real} equivalent, respectively for both BUP20 and SB20.
This is similar to the performance improvement we achieved through data augmentation \textit{Still-Virtual}.


Considerable performance improvements are obtained when using the \textit{virtual} sequences for training the spatio-temporal models.
The benefits of \textit{Temporal-Virtual} is clearly visible when comparing to the \textit{Still-Real} variants, with an absolute improvement of $4.6$ (BUP20) and $4.9$(SB20).
Furthermore, the \textit{Temporal-Virtual} system has a performance improvement of $3.7$ (BUP20) and $2.3$ (SB20) when compared to the \textit{Temporal-Real} system.
We attribute this increase in performance to the impact of using all images when calculating the loss compared to only the final image.
These results outline the benefit of the contextual information supplied from the temporal sequences for improved semantic segmentation allowing for temporal error reduction in pixel-wise classification.

\begin{table}[b!]
	\centering
	\caption{For a sequence containing 5 frames ($0\sim4$), same base model trained with \textit{virtual} outlining different performances for various number of frames.}
    \begin{tabular}{l l c  c}
    \toprule
    &\textbf{Model} & \textbf{BUP20}  & \textbf{SB20} \\\hline
    \midrule 
    1 & \textit{Temporal-Virtual} (N=5) & 81.9 & 76.2 \\
    2 & \textit{Temporal-Virtual} (N=4) & 81.3 & 75.7 \\
    3 & \textit{Temporal-Virtual} (N=3) & 81.2 & 74.9 \\
    4 & \textit{Temporal-Virtual} (N=2) & 80.9 & 74.0 \\\bottomrule
    \end{tabular}
    \label{tab:adapted_result}
    \vspace{-4mm}
\end{table}

To further evaluate the performance of our temporal network we explore the potential to augment the \textit{real} sequences by training on data from both directions, which led to a minor boost in performance of $79.5\%$ (BUP20) and $74.8\%$ (SB20) over the \textit{Temporal-Real}. 
Overall, these evaluations outline the importance of having fully annotated sequences as \textit{virtual} sequences outperforms the other experiments.

We also explore training a temporal model on \textit{virtual} sequences and then fine-tuning with \textit{real} sequences, referred to as \textit{Fine-Tuned} in line 3 of \tabref{tab:st_result}. 
This fine-tuning trick results in the highest mIoU score, out-performing both \textit{Temporal-Real} and \textit{Temporal-Virtual} models .
These results empirically support the use of \textit{virtual} sequences with convolutional re-sampling for training spatio-temporal models.

\subsection{Ablation Study; Varying the Number of Frames $N$ in a Temporal Sequence:}
\label{subsubsec:n_frames}

As an ablation study we explore the impact of varying the number of images in the temporal sequence.
We only consider a maximum case of five images, that is $N=\left[2,3,4,5\right]$, in alignment with previous experiments.
The results in~\tabref{tab:adapted_result} provide two important insights.
First, even for $N=2$ (the smallest sequence size) we achieve improved performance over using \textit{still} images.
This is an absolute performance improvement of $3.6$ (BUP20) and  $2.7$ (SB20) compared to the same \textit{Still-Real} systems in \tabref{tab:baseline_result}.
It indicates that even a small temporal sequence is able to provide more robust information to the classifier, resulting in more accurate segmentation.
Second, increasing the temporal field improves performance by incorporating previous predictions when performing pixel-wise classification.
In our experiments $N=5$ provided a suitable trade off between informative structure and temporal distance.
But future work can utilise larger GPU RAM to allow exploration into larger temporal sequences.


\section{Conclusion}
\label{sec:conc}

In this paper we presented a novel approach to augment temporal data for training agriculture based segmentation models.
By exploiting the relatively static scene and assuming that the robot (camera) moves we are able, with no extra annotation effort, to generate virtually labelled temporal sequences.
This is applied to sparsely annotated images within a video sequence to generate \textit{virtual} temporal sequences via structured crops and shifts.
We also explore the potential of a lightweight RNN (spatio-temporal model) by varying the ``feedback'' layers within the UNet architecture. 
It was found that introducing a convolutional layer as part of the ``feedback'' led to improved performance, which we attribute to being able to learn a representation over the feature map (when compared to a Bi-linear transform).
The validity of both our data augmentation technique and spatio-temporal (lightweight RNN) architecture was outlined by an increase in absolute performance of $4.6$ and $4.9$ when comparing two models, \textit{Still-Virtual} and \textit{Temporal-Virtual}, respectively for two datasets BUP20 and SB20.
These results outline the ability for our technique to work regardless of the properties of the scene being surveyed by the robotic platform, making it dataset agnostic.
We also showed empirically that increasing $N$, the number of frames, for our lightweight RNN increases performance, showing that temporal information does play a role in improving segmentation results.
Finally, this paper showed that an uncomplicated approach to generating temporally annotated data from sparsely annotated images improves segmentation performance. 

\section*{Acknowledgements}

This work was funded by the Deutsche Forschungsgemeinschaft (DFG, German Research Foundation) under Germany’s Excellence Strategy - EXC 2070 – 390732324.
%
%
\bibliographystyle{splncs04}
\bibliography{references}

\end{document}